%% file: main.tex
\documentclass[10pt, conference,letterpaper]{IEEEtran}
\makeatletter
\def\ps@headings{
\def\@oddhead{\mbox{}\scriptsize\rightmark \hfil 
}%
\def\@evenhead{\scriptsize
\hfil \leftmark\mbox{}}
\def\@oddfoot{}
\def\@evenfoot{}}
\usepackage{booktabs}
\makeatother 

\usepackage{colortbl}
\usepackage{xcolor}
\usepackage{array}

\usepackage{graphicx}
\usepackage{epstopdf}
\usepackage{cite} 
\usepackage{times}
\usepackage{dsfont} 
\usepackage{cite}
\usepackage{times}
\usepackage{pifont}
\usepackage{multirow}   
\usepackage{tikz}
\usetikzlibrary{shapes.geometric,calc}
   
\usepackage{amsthm}
\usepackage{graphicx}
\usepackage{subfigure}
\usepackage{color}
\usepackage{ifpdf}
\usepackage{epsfig}
\usepackage{latexsym}
\usepackage{amsfonts}
\usepackage{amssymb}
\usepackage{paralist}
\usepackage{comment}
\usepackage{xspace}
\usepackage{mathrsfs}
\usepackage{amssymb}
\usepackage{setspace}
\usepackage{color}
\usepackage[small]{caption}

\usepackage{subeqnarray}
\usepackage{amssymb}
\usepackage{url,epsfig,array}
\usepackage{leftidx}
\usepackage{amsmath}
\usepackage[T1]{fontenc}
\usepackage{aecompl}

\usepackage{calligra}
\usepackage{algorithm}
\usepackage{algorithmicx}
\usepackage{algpseudocode} 
\usepackage{amsmath}

\def\eg{\textit{e.g.}\xspace}

\makeatletter
\renewcommand{\maketag@@@}[1]{\hbox{\m@th\normalsize\normalfont#1}}%
\makeatother

\begin{document}
  
\title{\LARGE Adaptive and Fine-grained Module-wise Expert Pruning for Efficient LoRA-MoE Fine-Tuning }

\author{\IEEEauthorblockN{Weihang Li, Jianchun Liu, Hongli Xu}
\IEEEauthorblockA{  
$^1$School of Computer Science and Technology, University of Science and Technology of China, China\\
$^2$Suzhou Institute for Advanced Research, University of Science and Technology of China, China
} }

\maketitle

\begin{abstract}
\input{content/abstract.tex}
\end{abstract}
 

\section{Introduction}\label{sec:intro}
\input{content/intro.tex}

\section{Background and Motivation}\label{sec:motivation}
\input{content/motivation.tex}

\section{Proposed Framework}\label{sec:framework}
\input{content/framework.tex}

\section{Performance Evaluation}\label{sec:evaluation}
\input{content/evaluation.tex}

\section{Related Work}\label{sec:related}
\input{content/related.tex}

\section{Conclusion}\label{sec:conclusion}
\input{content/conclusion.tex}

\bibliographystyle{IEEEtran}
\bibliography{content/refs}

\end{document}

%% file: content/abstract.tex
LoRA-MoE has emerged as an effective paradigm for parameter-efficient fine-tuning, combining the low training cost of LoRA with the increased adaptation capacity of Mixture-of-Experts (MoE).
However, existing LoRA-MoE frameworks typically adopt a fixed and uniform expert configuration across heterogeneous Transformer modules (\eg, attention query/key projections and MLP gating networks), ignoring their distinct functional roles and capacity requirements.  
This design leads to localized over-provisioning, redundant trainable parameters, and unnecessary optimizer-state overhead. 
Moreover, prior methods enforce load balancing among experts throughout training. 
Although beneficial in the early stage, this constraint becomes restrictive once routing patterns stabilize, limiting expert specialization on downstream tasks.
In this paper, we propose DMEP, a novel LoRA-MoE fine-tuning framework based on Dynamic Module-wise Expert Pruning. 
DMEP tracks expert utilization during training and physically removes low-utility experts on a per-module basis, yielding a more compact expert structure tailored to different modules. 
The pruned model then continues training without the load-balancing constraint, freeing the remaining experts to focus entirely on the downstream task and develop specialized expertise. 
By jointly adapting module-wise expert capacity and eliminating unnecessary balancing, DMEP improves both parameter efficiency and training efficiency.
Extensive experiments on multiple reasoning benchmarks show that DMEP reduces trainable parameters by 35\%--43\% and improves training throughput by about 10\%, while maintaining or surpassing the downstream reasoning accuracy of uniform LoRA-MoE baselines.


%% file: content/intro.tex
In recent years, large language models (LLMs) have driven substantial progress across a wide range of natural language processing and multimodal tasks \cite{huang2026identifying}. Representative systems such as GPT-4 \cite{achiam2023gpt}, PaLM \cite{chowdhery2023palm}, and the LLaMA family \cite{touvron2023llama} learn rich semantic representations and world knowledge through self-supervised pre-training on massive text corpora. 
However, adapting these general-purpose foundation models to specialized downstream tasks via conventional full-parameter fine-tuning is often prohibitively expensive. 
For example, fully fine-tuning GPT-3 (175B) with the standard Adam optimizer requires about 1.2 TB of VRAM just to store model weights, gradients, and optimizer states, which makes independent deployment impractical \cite{hu2022lora}. 
To mitigate these limitations, Parameter-Efficient Fine-Tuning (PEFT) methods \cite{huo2026mitigating, lester2021power, li2021prefix, liu2025fedquad} have emerged as a practical alternative. 
Among them, Low-Rank Adaptation (LoRA) \cite{hu2022lora} stands out by decomposing task-specific weight updates into trainable low-rank matrices, while keeping the pre-trained backbone frozen.

Although LoRA attains competitive performance with only a small fraction of trainable parameters, its expressiveness is inherently limited by its static low-rank structure. 
For complex tasks that demand compositional reasoning or multi-step mathematical inference, a single low-rank projection often fails to model highly heterogeneous task-specific features \cite{qing2024alphalora, zhang2023adalora}. 
To address the capacity limitations of standard dense LoRA, recent studies have combined Mixture-of-Experts (MoE) \cite{jacobs1991adaptive, shazeer2017outrageously} with PEFT, which spurred the emergence of various LoRA-MoE frameworks, such as LoRAMoE \cite{dou2023loramoe}, MixLoRA \cite{li2024mixlora}, and MoELoRA \cite{liu2024moe}.
By replacing standard dense adapters with an ensemble of multiple specialized LoRA experts paired with a trainable gating network, LoRA-MoE models conditionally route input tokens to a sparse subset of experts. 
This paradigm significantly expands the overall learning capacity without proportionally increasing the active parameter count or per-token computational overhead. 

Nevertheless, existing LoRA-MoE methods typically adopt a fixed and uniform expert allocation strategy. 
That is, they assign the same number of experts across different Transformer layers, implicitly assuming that all parts of the network require similar representational capacity. 
This assumption is often misaligned with the actual behavior of downstream fine-tuning. 
Prior empirical studies on layer-wise MoE adaptation have shown that expert utilization can vary substantially across network depth \cite{aghdam2024moe, gao2024higher}. 
Shallow Transformer layers often focus on capturing more general lexical or syntactic patterns, which leads to concentrated routing behavior in which only a small subset of experts is frequently activated. 
In contrast, deeper layers are more responsible for resolving task-specific semantic divergence and high-level reasoning patterns, and therefore tend to benefit from broader expert participation. 
Under a rigid uniform allocation, this depth-dependent variation creates a clear capacity mismatch: \textit{some layers are over-provisioned with many experts that remain mostly idle, whereas others are constrained by insufficient expert capacity.}

To mitigate this redundancy, recent works have explored layer-wise expert allocation \cite{aghdam2024moe, gao2024higher, wang2025malora}, where different layers are assigned varied numbers of experts. 
While this design is more flexible than fully uniform allocation, it still operates at a relatively coarse granularity and overlooks a more subtle source of structural mismatch inside each Transformer layer. 
Specifically, different module types within the same Transformer layer perform fundamentally distinct computational roles, including attention projections (e.g., query, key, value, and output projections) and MLP projections (e.g., gate, up, and down projections). 
Attention projections mainly perform dynamic contextual token mixing, whereas MLP projections act more like point-wise knowledge transformations or memories \cite{geva2020transformer, dong2021attention}.
Because these modules process information in different ways, they can exhibit markedly different expert-utilization patterns even when they belong to the same layer.
For instance, during the fine-tuning of Qwen3-0.6B on the ScienceQA benchmark, we observe a clear utilization gap between attention and MLP projections in the 24th Transformer layer. 
The attention projection \texttt{O\_PROJ} shows highly skewed routing behavior, with its Gini coefficient\footnote{The Gini coefficient is computed over the expert routing-load distribution within a module. Given $E$ experts with routing loads $\{r_i\}_{i=1}^{E}$, it measures the inequality of expert utilization: a value close to 0 indicates nearly uniform routing across experts, while a larger value indicates that routing assignments are concentrated on a small subset of experts.} reaching 0.40. 
This indicates that only a small fraction of experts receive most token assignments, leaving many experts underutilized. 
In contrast, the adjacent MLP projection \texttt{GATE\_PROJ} maintains a Gini coefficient close to zero, suggesting a much flatter and more evenly distributed routing pattern. 
Therefore, \textit{assigning an identical number of experts to all modules within a layer still fails to match the actual module-level demand}, leaving substantial redundancy in some modules and insufficient capacity in others.

Beyond this structural rigidity, existing LoRA-MoE frameworks are also limited by a static training paradigm. 
In many cases, the expert architecture is determined before the main training process, often based on offline sensitivity analysis or manually selected hyperparameters \cite{zhang2023adalora}. 
However, routing behavior is not fixed: as fine-tuning progresses, the model continuously adapts to the downstream task, and expert preferences can evolve accordingly. 
An architecture determined offline may therefore become increasingly misaligned with the true online utilization pattern observed later in training.
Moreover, standard LoRA-MoE methods typically maintain an auxiliary load-balancing objective throughout the entire fine-tuning process. 
This regularization is important at the beginning of training because it prevents routing collapse and encourages all experts to receive sufficient gradient updates. 
Yet once expert usage patterns become relatively stable, persistent load balancing can become counterproductive.
At that stage, the task objective often favors more concentrated routing toward highly relevant experts, whereas the balancing objective continues to penalize such specialization and forces a more uniform workload distribution than the task itself requires. 
This tension can unnecessarily restrict the emergence of stronger task-specific expert specialization.

To address the above limitations, we propose \textbf{Dynamic Module-wise Expert Pruning (DMEP)}, a dynamic LoRA-MoE fine-tuning framework guided by online routing statistics. 
The key idea behind DMEP is that expert capacity should not be fixed in advance or shared uniformly across all modules. 
Instead, it should be progressively adapted according to the actual utilization patterns that emerge during downstream training. 
Concretely, DMEP begins from a load-aware dense expert initialization, which ensures sufficient expert exploration and stable optimization during the early training stage. 
After collecting reliable routing statistics, DMEP identifies low-utility experts at the module level and physically removes them from the model, directly reducing trainable parameters and optimizer-state overhead. 
Following this pruning stage, DMEP disables the load-balancing constraint and allows the remaining experts to focus entirely on the primary task objective. 
In this way, DMEP jointly addresses two inefficiencies of existing LoRA-MoE designs: redundant expert allocation caused by structurally mismatched capacity, and limited expert specialization caused by persistent balancing regularization.

Among the three stages of DMEP, the most critical design challenge lies in the pruning stage, where the framework must determine how many experts to retain for each functional module. 
This decision directly governs the trade-off between preserving adaptation capacity and reducing training overhead. 
If the retained expert budget is too small, the resulting module may suffer from insufficient expressive capacity, leading to degraded performance on complex downstream tasks. 
If the retained budget is too large, the framework removes too little redundancy and fails to deliver meaningful gains in parameter efficiency and throughput.
In summary, our main contributions are as follows:
\begin{itemize}
\item We identify two key inefficiencies in existing LoRA-MoE fine-tuning: coarse-grained expert allocation and persistent load balancing. Our analysis shows that expert utilization varies across both layers and intra-layer modules, making fixed or offline-determined allocation highly redundant.
  
    \item We propose DMEP, an adaptive LoRA-MoE fine-tuning framework that adapts module-wise expert capacity online using routing statistics. DMEP prunes low-utility experts and removes load balancing after routing stabilizes, reducing parameter overhead while enabling stronger task-specific specialization.
    \item Extensive experiments show that DMEP substantially reduces the total trainable parameters by 35\% to 43\% and improves training throughput by about 10\%, while maintaining competitive accuracy across multiple reasoning tasks.
\end{itemize}

%% file: content/motivation.tex
\subsection{LoRA-MoE Architecture}

Fine-tuning adapts pre-trained large language models (LLMs) to downstream tasks. However, updating billions of parameters incurs massive computational overhead and optimizer-state memory consumption \cite{liu2024hift,ma2026adafun}. For models at the LLaMA scale, this creates a strict bottleneck that makes rapid task adaptation computationally prohibitive \cite{touvron2023llama}.

Low-Rank Adaptation (LoRA) \cite{hu2022lora} mitigates this by freezing the pre-trained backbone and injecting trainable low-rank matrices. For a frozen pre-trained weight $\mathbf{W}_0 \in \mathbb{R}^{m \times n}$, LoRA introduces $\mathbf{A} \in \mathbb{R}^{r \times n}$ and $\mathbf{B} \in \mathbb{R}^{m \times r}$, where $r \ll \min(m,n)$. Only $\mathbf{A}$ and $\mathbf{B}$ are updated during training. The forward pass for an input $\mathbf{x}$ is defined as:
\begin{equation}
\mathbf{h} = \mathbf{W}_0 \mathbf{x} + \frac{\alpha}{r} \mathbf{B}\mathbf{A}\mathbf{x},
\end{equation}
where $\alpha$ is a scaling constant. By randomly initializing $\mathbf{A}$ and setting $\mathbf{B}$ to zero, the initial adaptation weight $\Delta \mathbf{W} = \mathbf{B}\mathbf{A}$ begins at $\mathbf{0}$. This efficiently reduces the trainable parameter count while preserving the original model knowledge.

However, LoRA's capacity is bounded by its low-rank constraint, limiting its ability to model complex task-specific patterns \cite{qing2024alphalora}. Simply increasing the rank $r$ scales memory and computational costs with diminishing returns. Combining LoRA with a Mixture-of-Experts (MoE) architecture \cite{shazeer2017outrageously} offers a scalable alternative. Recent work \cite{luo2024moelora, zadouri2023pushing} demonstrates that integrating multiple specialized experts and a trainable router improves model capacity without dramatically inflating the active parameter count per token.

Standard LoRA-MoE models replace specific Transformer linear layers—such as self-attention projections or feed-forward networks—with an ensemble of $N$ LoRA experts. The output is computed as:
\begin{equation}
\mathbf{o} = \mathbf{W}_0 \mathbf{x} + \sum_{i=1}^{N} g_i(\mathbf{x}) \, \mathbf{B}_i \mathbf{A}_i \mathbf{x},
\end{equation}
where $\mathbf{B}_i\mathbf{A}_i$ is the $i$-th expert and $g_i(\mathbf{x})$ is the routing probability. Typically, a Top-$k$ gating function activates only the highest-scoring experts per token. To prevent routing collapse, an auxiliary load-balancing loss is commonly added to the primary task loss:
\begin{equation}
L_{\text{total}} = L_{\text{task}} + \lambda L_{\text{aux}}.
\end{equation}
Although $L_{\text{aux}}$ ensures expert utilization during early training, strictly enforcing it throughout the entire optimization process, coupled with a rigid, uniform expert allocation, can severely constrain expert specialization. We detail these architectural and optimization bottlenecks below.

\subsection{Limitations of Existing Approaches}

Methods such as LoRAMoE \cite{dou2023loramoe}, MixLoRA \cite{li2024mixlora}, and Mixture-of-LoRAs \cite{feng2024mixture} uniformly distribute the same number of experts across all layers and modules. While this uniform design increases adaptation capacity, it can also introduce substantial parameter redundancy. To better understand this redundancy, we conduct a fine-grained routing analysis using a fully uniform configuration ($N=8$ experts per module) applied to Qwen3-0.6B\cite{yang2025qwen3} on the ScienceQA dataset\cite{lu2022learn}. We track the cumulative token-to-expert assignment frequencies across all targeted projections during the exploratory training phase to observe the actual expert utilization patterns.

\textbf{Quantitative Metrics.} To rigorously evaluate expert utilization and routing stability, we formalize three primary metrics at the module level based on our profiling data. Together, these metrics support the subsequent design of DMEP: the Gini coefficient measures module-level utilization imbalance, the routing entropy characterizes the sharpness of specialization, and the routing drift indicates when routing patterns become sufficiently stable for pruning. These metrics are not merely descriptive; they directly inform the subsequent design choices of DMEP, including how to identify redundant experts at the module level and when to execute pruning. First, we quantify the inequality of token allocation among experts within a given module $m$ using the module-wise \textbf{Gini Coefficient} ($G_m$):
\begin{equation}
    G_m = \frac{\sum_{i=1}^{N} (2i - N - 1) c_{m, (i)}}{N \sum_{i=1}^{N} c_{m, i}}
\end{equation}
where $c_{m, (i)}$ represents the cumulative discrete token load of the $i$-th expert in module $m$, sorted in non-decreasing order ($c_{m, (1)} \le c_{m, (2)} \le \dots \le c_{m, (N)}$). A value of $G_m=0$ indicates perfectly uniform routing, while $G_m \to 1$ reflects extreme load imbalance. 

Second, we compute the \textbf{Normalized Routing Entropy} ($H_m$) to evaluate the sharpness of routing decisions for module $m$:
\begin{equation}
    H_m = -\frac{1}{\log N} \sum_{i=1}^{N} p_{m, i} \log p_{m, i}
\end{equation}
where $p_{m, i} = \frac{c_{m, i}}{\sum_{j=1}^{N} c_{m, j}}$ is the normalized routing frequency derived from the discrete Top-$k$ token assignments. A higher entropy indicates a more uniform distribution (often forced by load-balancing penalties), whereas a lower entropy signifies sharper expert specialization. 

Third, to monitor the stability of these routing preferences over time, we track the module-wise \textbf{Routing Drift} ($D_m(t)$), defined as the $L_1$ distance between routing distributions at consecutive evaluation steps:
\begin{equation}
    D_m(t) = \left\| \mathbf{p}_m^{(t)} - \mathbf{p}_m^{(t-\Delta t)} \right\|_1
\end{equation}
where $\mathbf{p}_m^{(t)}$ is the routing distribution vector for module $m$ at training step $t$, and $\Delta t$ is a fixed logging interval (e.g., every 10 steps). A drift approaching $0$ indicates that the module's routing patterns have effectively converged.

\begin{figure}[t]
    \centering
    \includegraphics[width=0.85\linewidth]{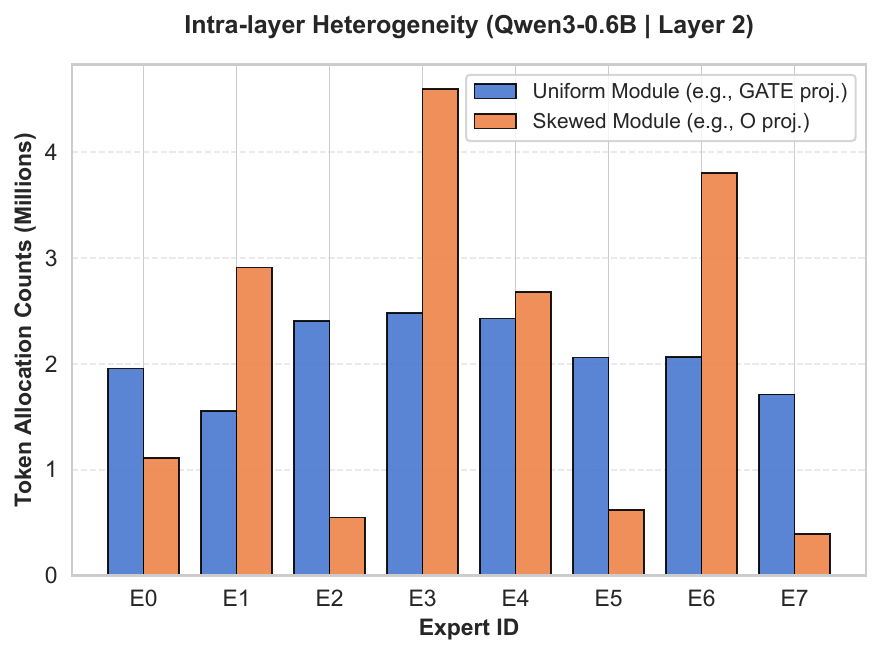}
    \caption{\textbf{Intra-layer heterogeneity of expert utilization (Qwen3-0.6B on ScienceQA).} The bar chart compares the token allocation across 8 experts for two adjacent modules in Layer 2. While the \texttt{GATE\_PROJ} module maintains a relatively uniform distribution, the \texttt{O\_PROJ} module exhibits significant skewness, where certain experts dominate the token traffic.}
    \label{fig:motivation}
\end{figure}

Our profiling reveals substantial heterogeneity in routing behavior. As illustrated in \textbf{Fig.~\ref{fig:motivation}}, even within the same Transformer block, different projection modules exhibit markedly different load patterns. For instance, in Layer 2, MLP modules (e.g., \texttt{GATE\_PROJ}) utilize experts relatively evenly, whereas attention projections (e.g., \texttt{O\_PROJ}) concentrate their workload on a few specific experts, leaving others virtually inactive. This example highlights pronounced intra-layer heterogeneity, showing that neighboring modules within the same layer can demand markedly different expert capacities. 

\begin{figure}[t]
    \centering
    \includegraphics[width=\linewidth]{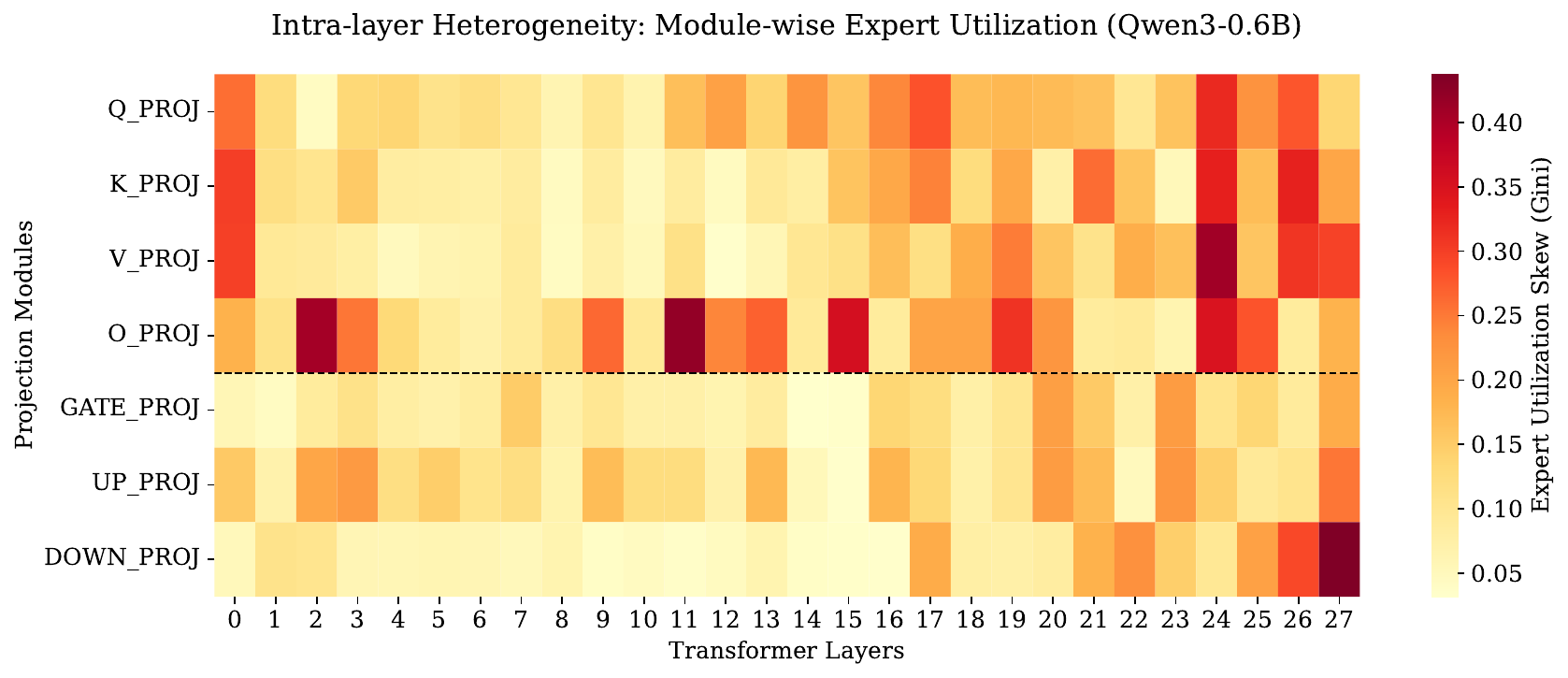}
    \caption{\textbf{Module-wise Expert Utilization Heatmap (Qwen3-0.6B).} The dashed line separates attention (top) and MLP (bottom) modules across 28 layers. The sharp vertical color contrast within the same layer—such as extreme load imbalance in \texttt{O\_PROJ} (dark red) versus uniform routing in \texttt{GATE\_PROJ} (light yellow)—reveals profound intra-layer heterogeneity. This confirms that uniform, layer-wise expert allocation ignores module-specific demands and causes severe capacity over-provisioning.}
    \label{fig:heatmap}
\end{figure}

\begin{figure}[t]
    \centering
    \includegraphics[width=0.9\linewidth]{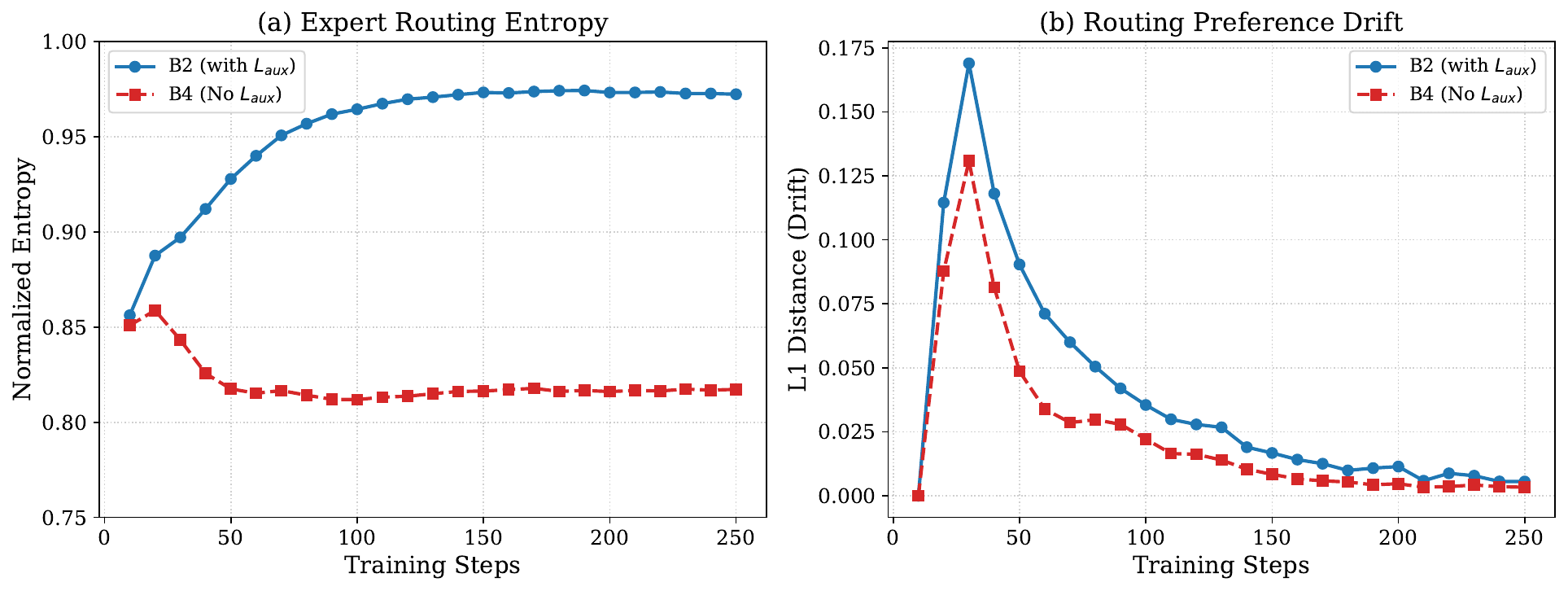}
    \caption{\textbf{Expert Routing Dynamics Analysis.} (a) Comparison of normalized routing entropy ($H_m$) between training with and without $L_{\text{aux}}$. (b) Evolution of routing drift ($D_m(t)$), showing rapid stabilization within the exploratory phase.}
    \label{fig:dynamics}
\end{figure}

\begin{figure}[t]
    \centering
    \includegraphics[width=0.7\linewidth]{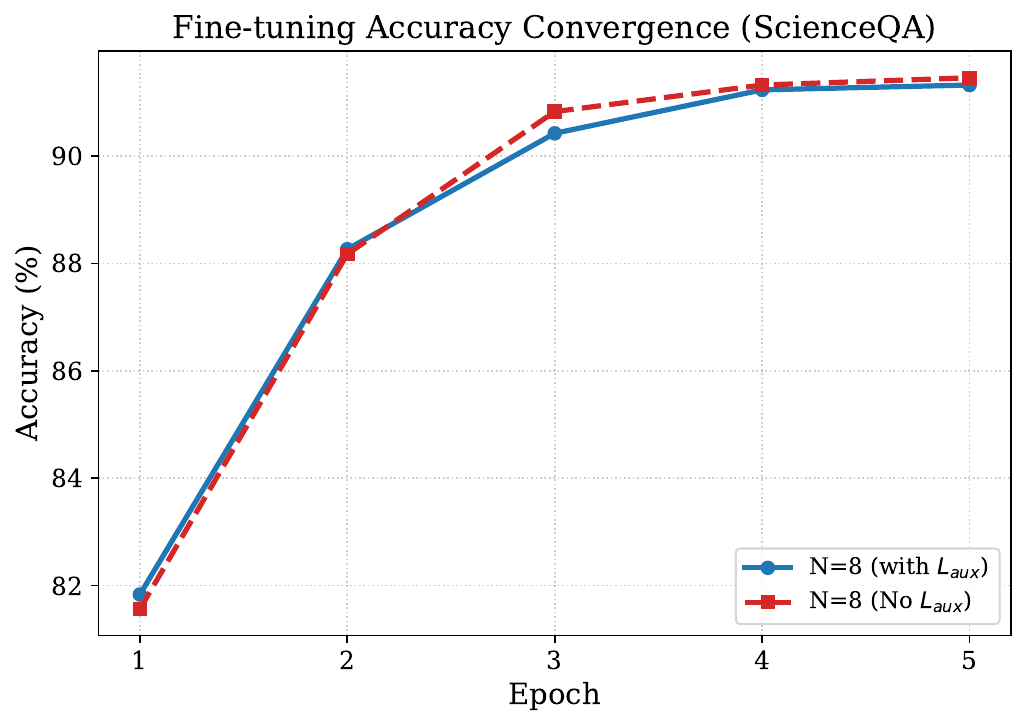}
    \caption{\textbf{Fine-tuning Accuracy Convergence on ScienceQA.} The curve demonstrates that training without $L_{\text{aux}}$ maintains stable convergence and achieves slightly higher peak accuracy compared to the balanced baseline.}
    \label{fig:accuracy}
\end{figure}

A closer examination reveals that existing static, coarse-grained heuristics lead to three major inefficiencies:

\begin{itemize}
    \item \textbf{Offline and Task-Agnostic Architecture Determination:} Existing non-uniform methods often rely on offline heuristics fixed before training. Because these static configurations are not learned from task-specific routing behavior, they often mismatch the actual module-level demand, leading to under-allocation in modules that require richer specialization.

    \item \textbf{Coarse-Grained Layer-Wise Allocation:} Current configurations typically assign an identical expert count to an entire Transformer layer, ignoring internal structural heterogeneity. As visualized in the heatmap (\textbf{Fig.~\ref{fig:heatmap}}), different attention and MLP modules can exhibit markedly different utilization patterns within the same layer. In particular, some attention projections, such as \texttt{O\_PROJ} and \texttt{K\_PROJ}, show substantially higher utilization inequality than adjacent MLP modules. This stark intra-layer variance implies that assigning a uniform, layer-wise expert count is too coarse and fails to accommodate the distinct representational demands of different architectural components.

    \item \textbf{Over-Regularization from Persistent Balancing:} Standard MoE training relies on an auxiliary load-balancing loss ($L_{\text{aux}}$) to prevent initial routing collapse. However, our profiling suggests that persistent $L_{\text{aux}}$ can constrain routing dynamics. As shown in \textbf{Fig.~\ref{fig:dynamics}(a)}, the configuration \textbf{with load-balancing loss} maintains a substantially higher routing entropy ($H_m \approx \textbf{0.972}$) than the configuration \textbf{without load-balancing loss} ($H_m \approx \textbf{0.817}$). Furthermore, \textbf{Fig.~\ref{fig:dynamics}(b)} indicates that the routing preference stabilizes with extremely low drift ($D_{\text{drift}} \approx \textbf{0.0355}$) exactly at step 100, which corresponds to the end of the dense exploratory phase. Crucially, as visualized in \textbf{Fig.~\ref{fig:accuracy}}, the setup \textbf{without load-balancing loss} achieves a slightly higher accuracy (\textbf{91.45\%}) compared to the balanced baseline (\textbf{91.32\%}). These observations indicate that persistent balancing may become unnecessary once useful routing patterns have emerged, and may even hinder sharper expert specialization. More importantly, they motivate disabling $L_{\text{aux}}$ after routing stabilization rather than enforcing it throughout the entire fine-tuning process.
\end{itemize}

Taken together, these observations motivate a dynamic framework that performs module-wise adaptive expert allocation and relaxes load-balancing regularization once routing patterns have stabilized.

%% file: content/framework.tex
\subsection{Overview}

\begin{figure*}[ht]
    \centering
    \includegraphics[width=0.95\linewidth]{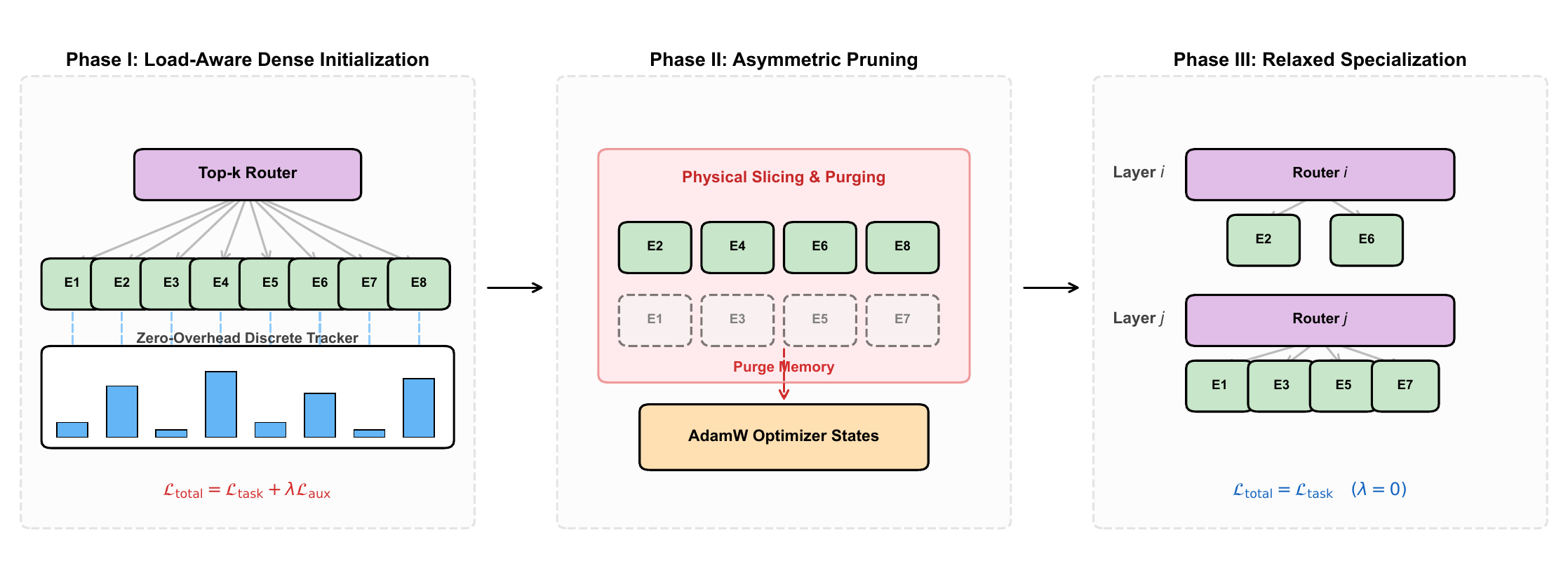}
    \caption{\textbf{The Overall Framework of Dynamic Module-wise Expert Pruning (DMEP).} \textbf{Phase I:} The model begins with a uniform dense initialization, where an online tracker records discrete routing frequencies. \textbf{Phase II:} Based on utilization scores, low-utility experts and their corresponding optimizer states are structurally removed from the model, eliminating redundant computation. \textbf{Phase III:} The model resumes fine-tuning with a heterogeneous architecture and disabled load-balancing loss, enabling task-driven specialization.}
    \label{fig:architecture}
\end{figure*}

To address the inherent rigidity of static expert allocation and the optimization conflicts introduced by persistent routing regularization, we propose \textbf{D}ynamic \textbf{M}odule-wise \textbf{E}xpert \textbf{P}runing (DMEP). DMEP is a dynamic, module-level optimization framework designed to discover a task-adaptive, fine-grained MoE architecture during the fine-tuning process. Instead of relying on manual heuristics, DMEP transitions the network from a dense, uniform initialization to an efficient, heterogeneous structure through three sequential phases, as illustrated in \textbf{Fig.~\ref{fig:architecture}}:

\begin{itemize}
    \item \textbf{Phase I: Load-Aware Dense Initialization.} The model begins with a uniform dense initialization. During this exploratory stage, an online tracker records discrete routing frequencies with negligible additional overhead, while an auxiliary load-balancing loss prevents early routing collapse.
    
    \item \textbf{Phase II: Fine-Grained Module-Wise Pruning.} Once initial routing patterns stabilize, we execute a structural, module-wise physical pruning step. Based on empirical utilization scores, redundant experts and their corresponding optimizer states are structurally removed from the model, yielding a heterogeneous architecture.
    
    \item \textbf{Phase III: Relaxed Specialization.} The pruned network resumes fine-tuning. At this stage, we completely disable the auxiliary routing regularization ($\lambda = 0$), eliminating optimization conflicts and enabling more task-driven expert specialization.
\end{itemize}

This phased design follows directly from the observations in the previous section: module-wise routing heterogeneity motivates structural pruning at the module level, while routing stabilization justifies disabling load balancing after the exploratory phase. Before detailing the specific implementation mechanics of these three phases, we first formalize the mathematical foundation of uniform LoRA-MoE.

\subsection{Preliminaries: Uniform LoRA-MoE}
Consider a pre-trained Transformer model where specific dense linear projections (e.g., query, key, value, and feed-forward up/down projections) are targeted for adaptation. Let $\mathbf{W}_0 \in \mathbb{R}^{d_{\text{in}} \times d_{\text{out}}}$ be the frozen pre-trained weight matrix of a targeted module $m$. In a standard LoRA-MoE architecture, this dense module is augmented with $N$ low-rank experts. The $i$-th expert consists of a down-projection $\mathbf{A}_{m,i} \in \mathbb{R}^{d_{\text{in}} \times r}$ and an up-projection $\mathbf{B}_{m,i} \in \mathbb{R}^{r \times d_{\text{out}}}$, where $r \ll \min(d_{\text{in}}, d_{\text{out}})$ is the intrinsic rank. 

For an input token $\mathbf{x} \in \mathbb{R}^{d_{\text{in}}}$, the routing network computes the assignment probabilities $\mathbf{p}_m(\mathbf{x}) = \text{Softmax}(\mathbf{W}_{g,m} \mathbf{x})$, where $\mathbf{W}_{g,m} \in \mathbb{R}^{d_{\text{in}} \times N}$ is the gating weight. The forward pass of the uniform LoRA-MoE for module $m$ is defined as:
\begin{equation}
    \mathbf{h}_m = \mathbf{W}_0 \mathbf{x} + \alpha \sum_{i \in \mathcal{S}(\mathbf{x})} p_{m,i}(\mathbf{x}) \cdot (\mathbf{B}_{m,i} \mathbf{A}_{m,i} \mathbf{x})
\end{equation}
where $\alpha$ is a constant scaling factor, and $\mathcal{S}(\mathbf{x})$ denotes the set of indices selected by the Top-$k$ routing mechanism (i.e., the $k$ experts with the highest probabilities).

To prevent routing collapse during the initial training steps, a standard auxiliary load-balancing loss $L_{\text{aux}}$ is typically employed:
\begin{equation}
    L_{\text{aux}} = N \sum_{i=1}^{N} f_i \cdot P_i
\end{equation}
where $f_i$ is the fraction of tokens assigned to expert $i$ within a batch, and $P_i$ is the mean routing probability for expert $i$ across the batch. While $L_{\text{aux}}$ ensures that all experts receive gradients initially, continuously enforcing it tends to encourage a more uniform routing distribution, which contradicts the goal of specialized, heterogeneous capacity allocation.

\subsection{Phase I: Load-Aware Dense Initialization}
Unlike existing non-uniform methods that impose a fixed architectural prior before training begins, DMEP adopts a data-driven discovery process. We start by initializing a fully uniform LoRA-MoE architecture, allocating $N$ experts to every targeted projection across all layers. 

During the exploratory phase (e.g., the first training epoch, $E_w=1$), we maintain the auxiliary load-balancing loss ($L_{\text{aux}}$) alongside the primary task objective. By setting the coefficient $\lambda > 0$, the optimization objective becomes $L_{\text{total}} = L_{\text{task}} + \lambda L_{\text{aux}}$. This regularization encourages broader exploration of the initialized expert space, helping establish initial routing preferences without severe early gradient starvation.

To capture the true expert demand, DMEP employs an online tracking mechanism. Unlike previous works that aggregate continuous Softmax probabilities, DMEP tracks the exact hard routing decisions. For each input token, the cumulative discrete load $U_{m,i}$ for the $i$-th expert in module $m$ over the exploratory phase is accumulated as:
\begin{equation}
    U_{m, i} = \sum_{t=1}^{T_{\text{explore}}} \sum_{\mathbf{x} \in \mathbf{X}^{(t)}} \mathbb{I} \big( i \in \text{TopK}(\mathbf{p}_m(\mathbf{x}), k) \big)
\end{equation}
where $\mathbb{I}(\cdot)$ is the indicator function and $\mathbf{X}^{(t)}$ is the batch of tokens at step $t$. Using discrete Top-$k$ assignments makes the utilization statistics more consistent with the actual routing decisions that determine expert activation during training. 

From a systems perspective, this tracking mechanism introduces negligible additional overhead. The accumulation only requires counting routed expert indices, which can be implemented with lightweight non-differentiable tensor operations inside a \texttt{no\_grad} context. At the end of Phase I, we normalize the counts to obtain the relative utilization score $s_{m, i} = U_{m, i} / \sum_{j=1}^{N} U_{m, j}$, which serves as the empirical basis for architectural pruning.

\subsection{Phase II: Fine-Grained Module-Wise Pruning}
Once routing drift becomes sufficiently small and routing preferences stabilize, DMEP executes a one-shot, fine-grained pruning step. We adopt a one-shot pruning strategy to avoid repeated architecture mutation and optimizer-state reconstruction, which would otherwise introduce additional instability and systems overhead. We introduce a global threshold $\tau \in (0, 1)$ to filter redundant capacity. An expert $i$ is slated for removal if $s_{m, i} < \tau$. To prevent complete module collapse and maintain baseline representation power, we enforce a minimum survival constraint $K_{\min} \ge k$. This constraint ensures that Top-$k$ routing remains well-defined after pruning and prevents over-pruning from collapsing a module into an under-capacity bottleneck. The final survivor set $\mathcal{A}_m$ is determined as:
\begin{equation}\label{eq:survivor_set}
\mathcal{A}_m = 
\begin{cases} 
\{ i \mid s_{m, i} \ge \tau \}, & \text{if } \sum_{i=1}^{N} \mathbb{I}(s_{m, i} \ge \tau) \ge K_{\min}, \\
\text{TopK}_{\mathrm{idx}}(\mathbf{s}_m, K_{\min}), & \text{otherwise}.
\end{cases}
\end{equation}

\textbf{Physical Parameter and Optimizer State Slicing.} Conventional pruning techniques often apply binary zero-masks to the weights. While masking prevents pruned parameters from influencing the forward pass, the model still performs dense matrix multiplications with zeros. More critically, popular adaptive optimizers like AdamW continue to update the first moment ($\mathbf{m}^{(t)}$) and second moment ($\mathbf{v}^{(t)}$) for these zero-masked parameters, continuing to consume unnecessary optimizer memory and update cost.

To achieve true structural sparsity and throughput acceleration, DMEP performs \textit{structural physical slicing}. Let $\Theta_{m} \in \mathbb{R}^{N \times d}$ denote a stacked parameter tensor for module $m$. DMEP explicitly reconstructs the weight tensors by indexing only the surviving dimensions:
\begin{equation}
    \Theta_{m}^{\text{pruned}} = \Theta_{m}[\mathcal{A}_m, :]
\end{equation}
Simultaneously, DMEP integrates deeply with the optimizer graph. It locates the removed parameter blocks within the optimizer's state dictionary and physically purges their corresponding momentum and variance tensors:
\begin{equation}
    \mathbf{m}^{\text{pruned}} = \mathbf{m}[\mathcal{A}_m, :], \quad \mathbf{v}^{\text{pruned}} = \mathbf{v}[\mathcal{A}_m, :]
\end{equation}
Consequently, pruning a single expert permanently removes it from both the forward computation graph and the optimizer update step. This structural removal ensures that subsequent training focuses only on the surviving parameters, reducing redundant overhead. The gating network $\mathbf{W}_{g,m}$ is synchronously re-indexed to output dimensions matching $|\mathcal{A}_m|$. Therefore, DMEP translates routing sparsity into real structural compression, instead of simply masking inactive experts with zeros.

\subsection{Phase III: Relaxed Specialization}
Following the structural pruning, the model enters Phase III with a fixed, highly heterogeneous architecture adapted to the downstream task. At this juncture, we completely disable the auxiliary load-balancing loss by setting its coefficient to zero ($\lambda = 0$). The training objective simplifies to:
\begin{equation}
    L_{\text{total}} = L_{\text{task}}
\end{equation}

This relaxation changes the role of routing regularization in MoE fine-tuning. We hypothesize that the main routing preferences are established during the exploratory phase. Once these preferences stabilize, the necessity of $L_{\text{aux}}$ diminishes. Keeping $L_{\text{aux}}$ active may conflict with the primary task loss, artificially elevating routing entropy and blurring expert focus.

By eliminating this penalty, Phase III yields a less constrained optimization setting. The surviving, highly utilized experts are free to specialize further under the primary task objective. The routing mechanism is no longer penalized for concentrating tokens on a narrow subset of highly specialized experts, allowing the heterogeneous architecture to develop sharper task-specific specialization. This design is consistent with the routing analysis in the previous section, which shows that persistent balancing becomes less beneficial once routing preferences have stabilized.

\subsection{Algorithm Summary}
The overall training procedure of DMEP is summarized in Algorithm~\ref{alg:dmep}. It consists of load-aware dense exploration, one-shot module-wise pruning, and continued fine-tuning on the resulting heterogeneous architecture.
\begin{algorithm}[ht]
\caption{Dynamic Module-wise Expert Pruning (DMEP) for LoRA-MoE}
\label{alg:dmep}
\textbf{Require:} Pre-trained LLM backbone, training dataset $\mathcal{D}$, total fine-tuning epochs $E$. \\
\textbf{Hyperparameters:} Warm-up epoch $E_w = 1$, pruning threshold $\tau$, minimum surviving experts $K_{\min}$.
\begin{algorithmic}[1]
\State Initialize a uniform LoRA-MoE architecture with $N$ experts per module.
\State Set the auxiliary load-balancing coefficient $\lambda > 0$.
\State Initialize cumulative load $U_{m,i} \gets 0$ for all modules $m$ and experts $i$.
\vspace{0.15cm}
\State \textbf{\# Phase I: Load-Aware Dense Initialization}
\For{epoch $e = 1, 2, \dots, E_w$}
    \For{each batch $\mathbf{X} \in \mathcal{D}$}
        \State Compute module-wise router probabilities $\mathbf{p}_m(\mathbf{x})$ and execute Top-$k$ routing.
        \State Update model parameters by minimizing $L_{\text{task}} + \lambda L_{\text{aux}}$.
        \State Accumulate discrete token routing frequencies:
        \State \quad $U_{m,i} \gets U_{m,i} + \sum_{\mathbf{x} \in \mathbf{X}} \mathbb{I} \big( i \in \text{TopK}(\mathbf{p}_m(\mathbf{x}), k) \big)$.
    \EndFor
\EndFor
\vspace{0.15cm}
\State \textbf{\# Phase II: Fine-Grained Module-Wise Pruning}
\For{each targeted module $m$}
    \State Compute relative utilization scores $s_{m,i} = U_{m,i} / \sum_{j=1}^{N} U_{m,j}$.
    \State Determine the surviving expert set $\mathcal{A}_m$ based on $\tau$ and $K_{\min}$ (Eq.~\ref{eq:survivor_set}).
    \State \textbf{Structurally remove} the parameters of pruned experts ($i \notin \mathcal{A}_m$) from the model.
    \State \textbf{Purge} the corresponding optimizer states (e.g., momentum $\mathbf{m}$, variance $\mathbf{v}$).
    \State Re-index the gating network output to match the surviving expert set $\mathcal{A}_m$.
\EndFor
\vspace{0.15cm}
\State \textbf{\# Phase III: Relaxed Specialization}
\State Disable auxiliary routing regularization: $\lambda \gets 0$.
\For{epoch $e = E_w + 1, \dots, E$}
    \For{each batch $\mathbf{X} \in \mathcal{D}$}
        \State Perform forward passes over the pruned, heterogeneous architecture.
        \State Update the surviving parameters by minimizing only $L_{\text{task}}$.
    \EndFor
\EndFor
\vspace{0.15cm}
\State \Return The pruned and fine-tuned heterogeneous LoRA-MoE model.
\end{algorithmic}
\end{algorithm}

\subsection{Complexity and Efficiency Analysis}
The transition from a uniform to a heterogeneous architecture alters the computational footprint of the model. Let $L$ be the number of Transformer layers, $M$ be the number of targeted modules per layer, and $N$ be the initial number of experts per module. The total trainable parameter count during Phase I is $\mathcal{O}(L \cdot M \cdot N \cdot r \cdot (d_{\text{in}} + d_{\text{out}}))$.

After the Phase II pruning, the number of surviving experts in module $m$ of layer $l$ becomes $N_{l,m} \le N$. The total retained parameter complexity shrinks to $\mathcal{O}(r \cdot (d_{\text{in}} + d_{\text{out}}) \sum_{l=1}^{L} \sum_{m=1}^{M} N_{l,m})$. The physical slicing ensures that the optimizer no longer maintains or updates states for pruned experts, reducing the computation and memory cost of the backward pass and optimizer update. Furthermore, because the gating network now computes logits over a smaller average expert set $\bar{N} \ll N$, the routing overhead is correspondingly reduced, translating to higher training throughput in Phase III. This reduction in retained parameters, optimizer-state updates, and routing overhead is consistent with the throughput improvements reported in the experimental results.

%% file: content/evaluation.tex
To rigorously evaluate the effectiveness of Dynamic Module-wise Expert Pruning (DMEP), we conduct fine-tuning experiments focusing on both adaptation performance and computational efficiency. 

\subsection{Experimental Setup}

\textbf{Hyperparameters and Training Details.} For all fine-tuning configurations, the intrinsic LoRA rank is set to $r=8$ with a scaling factor of $\alpha=16$. The adapters are applied to all linear projections within the self-attention and feed-forward networks (i.e., \texttt{q\_proj}, \texttt{k\_proj}, \texttt{v\_proj}, \texttt{o\_proj}, \texttt{gate\_proj}, \texttt{up\_proj}, \texttt{down\_proj}), resulting in 7 targeted modules per Transformer layer. For MoE routing, we employ a Top-$2$ assignment strategy ($k=2$). All models are fine-tuned for 5 epochs using \texttt{bfloat16} mixed precision. We optimize the parameters using AdamW with a peak learning rate of $3 \times 10^{-4}$ and a cosine decay schedule following a 10\% linear warm-up phase. To ensure fair optimization dynamics across models of vastly different sizes, we maintain a strict effective batch size of 128. For the 0.6B model, this is achieved using a per-device batch size of 16 and 8 gradient accumulation steps, whereas for the memory-intensive 8B model, we reduce the per-device batch size to 4 and increase gradient accumulation to 32 steps \cite{ma2025asynchronous}. 

The maximum sequence length is set to 256, and all inputs are formatted using a unified instruction prompt paired with the model's default tokenizer. For DMEP configurations, the exploratory warm-up epoch is set to $E_w=1$, and the minimum survival constraint is $K_{\min}=2$. Unless otherwise specified, the same DMEP hyperparameters are used across all tasks and model scales, and all experiments are conducted with a fixed random seed (42) to ensure reproducibility.

\textbf{Models.} To evaluate the scalability and generalizability of our method across different model capacities, we utilize two base models from the Qwen3 family: the lightweight \textbf{Qwen3-0.6B} for extensive architectural ablation and baseline comparisons, and the larger \textbf{Qwen3-8B}\cite{yang2025qwen3} to demonstrate the framework's effectiveness on state-of-the-art scale models.

\textbf{Datasets and Preprocessing.} We evaluate our method on three representative reasoning tasks:
\begin{itemize}
    \item \textbf{ScienceQA \cite{lu2022learn}:} A multi-modal science question-answering benchmark. We isolate the text-based reasoning capabilities by filtering out questions requiring visual context, formulating the task as a standard multiple-choice format.
    \item \textbf{OpenBookQA \cite{mihaylov2018can}:} A question-answering dataset requiring common sense and elementary science knowledge. We preprocess this into a strict 4-way multiple-choice format (A/B/C/D).
    \item \textbf{GSM8K \cite{cobbe2021training}:} We cast GSM8K as direct answer prediction without chain-of-thought generation. The model is trained to produce the final numeric answer after a fixed delimiter (\texttt{\#\#\#\#}), and evaluation is performed through exact answer extraction with regular expressions.
\end{itemize}

\textbf{Hardware Infrastructure.} All experiments are conducted on a high-performance workstation using a single NVIDIA RTX A6000 GPU (48GB VRAM) and an Intel Xeon Platinum 8358P CPU. The software environment is built upon PyTorch 2.6.0 and CUDA 12.4.

\textbf{Baselines and Configurations.} We compare DMEP against representative adaptation paradigms to establish clear performance boundaries:
\begin{itemize}
    \item \textbf{Dense LoRA:} Standard Low-Rank Adaptation without routing mechanisms ($N=1, k=1$). This serves as a lightweight adaptation baseline.
    \item \textbf{Symmetric MoE ($N=8$):} A symmetric Mixture-of-Experts configuration allocating 8 experts uniformly across all modules with load-balancing active ($N=8, k=2, \lambda > 0$). This serves as a high-capacity MoE baseline.
    \item \textbf{DMEP ($\tau=0.10$):} Our proposed Dynamic Module-wise Expert Pruning framework. We initialize the model symmetrically ($N=8, k=2$) and perform pruning at the default threshold of $\tau=0.10$, enabling relaxed specialization in subsequent epochs.
\end{itemize}

\textbf{Evaluation Metrics.} To provide a holistic assessment of both task performance and system efficiency, we track three key metrics: (1) \textbf{Accuracy}, measuring the correct prediction rate on the test set; (2) \textbf{Trainable Parameters}, quantifying the active parameter count before and after pruning; and (3) \textbf{Training Throughput}, measured in tokens processed per second (tok/s) to evaluate the actual computational speedup yielded by structural pruning.

\subsection{Experimental Results}

The performance of DMEP compared to the baselines across various model scales and datasets is summarized in Table~\ref{tab:main_results}. Overall, the data shows that DMEP consistently improves parameter efficiency and post-pruning throughput relative to the symmetric MoE baseline while preserving comparable accuracy in most settings.

\begin{table*}[ht]
\centering
\caption{\textbf{Main Results on Fine-tuning Performance and Efficiency.} We compare Dense LoRA, Symmetric MoE ($N=8$), and AEP ($\tau=0.10$) across three reasoning tasks. "Params" indicates the active trainable parameter count. "Throughput" indicates the processing speed (tokens per second) during the final training phase. Bold numbers indicate the leading performance among MoE-based architectures.}
\label{tab:main_results}
\resizebox{\textwidth}{!}{
\begin{tabular}{lllrcc}
\toprule
\textbf{Base Model} & \textbf{Dataset} & \textbf{Method} & \textbf{Params (M)} & \textbf{Throughput (tok/s)} & \textbf{Accuracy (\%)} \\
\midrule
\multirow{9}{*}{\textbf{Qwen3-0.6B}} & \multirow{3}{*}{ScienceQA} 
& Dense LoRA & 5.0 & 2620 & 89.03 \\
&& Symmetric MoE ($N=8$) & 42.7 & 1368 & 91.32 \\
&& \cellcolor{gray!15}\textbf{AEP (Ours)} & \cellcolor{gray!15}\textbf{24.2} & \cellcolor{gray!15}\textbf{1466} & \cellcolor{gray!15}\textbf{91.55} \\
\cmidrule{2-6}
& \multirow{3}{*}{OpenBookQA} 
& Dense LoRA & 5.0 & 1753 & 66.00 \\
&& Symmetric MoE ($N=8$) & 42.7 & 923 & \textbf{69.00} \\
&& \cellcolor{gray!15}\textbf{AEP (Ours)} & \cellcolor{gray!15}\textbf{24.2} & \cellcolor{gray!15}\textbf{1032} & \cellcolor{gray!15}\textbf{69.00} \\
\cmidrule{2-6}
& \multirow{3}{*}{GSM8K} 
& Dense LoRA & 5.0 & 2563 & 13.00 \\
&& Symmetric MoE ($N=8$) & 42.7 & 1354 & 17.00 \\
&& \cellcolor{gray!15}\textbf{AEP (Ours)} & \cellcolor{gray!15}\textbf{25.3} & \cellcolor{gray!15}\textbf{1502} & \cellcolor{gray!15}\textbf{19.00} \\
\midrule
\multirow{9}{*}{\textbf{Qwen3-8B}} & \multirow{3}{*}{ScienceQA} 
& Dense LoRA & 21.8 & 526 & 94.96 \\
&& Symmetric MoE ($N=8$) & 185.2 & 294 & \textbf{97.08} \\
&& \cellcolor{gray!15}\textbf{AEP (Ours)} & \cellcolor{gray!15}\textbf{121.8} & \cellcolor{gray!15}\textbf{320} & \cellcolor{gray!15}96.40 \\
\cmidrule{2-6}
& \multirow{3}{*}{OpenBookQA} 
& Dense LoRA & 21.8 & 353 & 93.00 \\
&& Symmetric MoE ($N=8$) & 185.2 & 196 & \textbf{95.00} \\
&& \cellcolor{gray!15}\textbf{AEP (Ours)} & \cellcolor{gray!15}\textbf{107.9} & \cellcolor{gray!15}\textbf{216} & \cellcolor{gray!15}\textbf{95.00} \\
\cmidrule{2-6}
& \multirow{3}{*}{GSM8K} 
& Dense LoRA & 21.8 & 517 & 42.00 \\
&& Symmetric MoE ($N=8$) & 185.2 & 288 & \textbf{43.00} \\
&& \cellcolor{gray!15}\textbf{AEP (Ours)} & \cellcolor{gray!15}\textbf{110.4} & \cellcolor{gray!15}\textbf{317} & \cellcolor{gray!15}\textbf{43.00} \\
\bottomrule
\end{tabular}
}
\end{table*}

\textbf{Task Performance.} 
The experimental results indicate that DMEP maintains competitive representational capacity while achieving significant parameter reductions. For example, on Qwen3-8B with OpenBookQA, DMEP reduces the active trainable parameter count from 185.2M to 107.9M, a reduction of approximately 41.7\%, while preserving the same peak accuracy of 95.00\%. On Qwen3-0.6B with GSM8K, DMEP reduces the parameter count by 40.7\% (42.7M to 25.3M)  and simultaneously improves accuracy from 17.00\% to 19.00\%. These observations are consistent with the design motivation of DMEP: once routing patterns have been established during the exploratory phase, removing persistent load-balancing constraints allows the surviving experts to specialize more effectively.

\textbf{Effect of Model Scale.}
An interesting trend is that the impact of DMEP varies across model scales. On the smaller Qwen3-0.6B model, DMEP not only reduces parameter redundancy but also yields modest accuracy gains on ScienceQA and GSM8K. In contrast, on Qwen3-8B, the primary advantage of DMEP lies in improving parameter efficiency and throughput while largely preserving the strong performance of the symmetric baseline. This suggests that the proposed pruning strategy is particularly beneficial for smaller models with limited effective capacity, whereas for larger models, it mainly serves as an efficiency-oriented compression mechanism.

\textbf{Effect across Tasks.}
The impact of DMEP also shows variations across task types. On ScienceQA and OpenBookQA, DMEP generally matches the symmetric MoE baseline while using substantially fewer parameters. On the more challenging GSM8K task, DMEP yields a larger relative improvement over both Dense LoRA and symmetric MoE in the 0.6B scale. This pattern suggests that more challenging reasoning tasks may benefit more from the concentrated capacity induced by dynamic pruning.

\textbf{Limitations of Pruning.}
DMEP does not uniformly outperform the fully symmetric baseline in every setting. For example, on Qwen3-8B with ScienceQA, the symmetric MoE still achieves the highest accuracy (97.08\% vs. 96.40\%). This observation suggests that for tasks where the full expert budget is already effectively utilized during exploration, pruning may introduce a small loss in representational flexibility. Nevertheless, this performance gap remains small relative to the substantial gains in parameter efficiency and the consistent throughput improvements observed after pruning.

\textbf{Accuracy--Efficiency Trade-off.}
Taken together, the results indicate that DMEP shifts the accuracy--efficiency trade-off to a more favorable regime. Compared with the symmetric MoE baseline, DMEP consistently reduces the active parameter count and improves training throughput, while preserving comparable accuracy in most settings and even improving it in several cases. Relative to Dense LoRA, DMEP retains the representational advantage of multi-expert adaptation while avoiding a large portion of the redundancy inherent in a fully symmetric MoE design.

\subsection{Ablation Study}

To systematically isolate the contributions of our proposed mechanisms and understand the sensitivity of DMEP to its hyperparameters, we conduct extensive ablation studies using the Qwen3-0.6B model on the ScienceQA dataset. The results are summarized in Table~\ref{tab:ablation_results}.

\begin{table*}[ht]
\centering
\caption{\textbf{Ablation Study Results.} All experiments are conducted on Qwen3-0.6B with the ScienceQA dataset. We analyze the impact of forced symmetric allocation, pruning thresholds ($\tau$), and pruning timing (measured in steps). "Params" indicates the active trainable parameters. Throughput is recorded during the final training phase.}
\label{tab:ablation_results}
\setlength{\tabcolsep}{24pt} 
\begin{tabular}{lcccc}
\toprule
\textbf{Configuration} & \textbf{Threshold} & \textbf{Params} & \textbf{Throughput} & \textbf{Accuracy} \\
\midrule
\multicolumn{5}{l}{\textit{Symmetric Baselines}} \\
Symmetric MoE ($N=8$) & - & 42.7M & 1368 tok/s & 91.32\% \\
Symmetric MoE ($N=4$) & - & 21.3M & 1439 tok/s & 90.29\% \\
\midrule
\multicolumn{5}{l}{\textit{DMEP Pruning Thresholds (Pruned at Step 100)}} \\
DMEP (Conservative) & $\tau=0.05$ & 39.7M & 1410 tok/s & 91.37\% \\
\cellcolor{gray!15}\textbf{DMEP (Default)} & \cellcolor{gray!15}\textbf{$\tau=0.10$} & \cellcolor{gray!15}\textbf{31.2M} & \cellcolor{gray!15}\textbf{1438 tok/s} & \cellcolor{gray!15}\textbf{91.55\%} \\
DMEP (Aggressive) & $\tau=0.15$ & 13.1M & 1508 tok/s & 90.65\% \\
DMEP (Extreme) & $\tau=0.20$ & 10.8M & 1511 tok/s & 90.74\% \\
\midrule
\multicolumn{5}{l}{\textit{DMEP Pruning Timing (Threshold $\tau=0.10$)}} \\
Early Pruning & Step 50 & 26.7M & 1457 tok/s & 91.23\% \\
\cellcolor{gray!15}\textbf{Optimal Pruning} & \cellcolor{gray!15}\textbf{Step 100} & \cellcolor{gray!15}\textbf{31.2M} & \cellcolor{gray!15}\textbf{1438 tok/s} & \cellcolor{gray!15}\textbf{91.55\%} \\
Late Pruning & Step 250 & 35.3M & 1275 tok/s & 91.23\% \\
\bottomrule
\end{tabular}
\end{table*}

\textbf{Dynamic vs. Symmetric Capacity Reduction.} 
A core motivation of DMEP is that dynamically allocating experts based on actual routing demand can be more effective than enforcing a uniform symmetric bottleneck. To validate this, we compare the aggressive DMEP variant ($\tau=0.15$) with a constrained symmetric baseline ($N=4$). The forced symmetric $N=4$ configuration retains 21.3M parameters and achieves an accuracy of 90.29\%. In contrast, DMEP ($\tau=0.15$) actively prunes the network to an asymmetric structure containing only 13.1M parameters (a $\sim$38\% reduction compared to $N=4$), yet it achieves a higher accuracy of 90.65\% and faster throughput (1508 tok/s vs. 1439 tok/s). This comparison suggests that uniformly shrinking expert capacity can under-provision certain important modules, whereas dynamic pruning preserves capacity in a more targeted manner.

\textbf{Impact of the Pruning Threshold ($\tau$).} 
The threshold $\tau$ dictates the trade-off between model capacity and computational efficiency. As shown in Table~\ref{tab:ablation_results}, a conservative threshold ($\tau=0.05$) prunes the network lightly (39.7M params) and maintains a high accuracy of 91.37\%. Increasing the threshold to the default $\tau=0.10$ provides the best balance among the tested settings, achieving the highest accuracy of 91.55\% with 31.2M active parameters. Further increasing $\tau$ to 0.15 and 0.20 induces higher structural sparsity, reducing the parameter footprint to 13.1M and 10.8M, respectively. While this aggressive pruning pushes throughput beyond 1500 tok/s, it incurs a slight accuracy degradation. These results indicate a clear accuracy--efficiency trade-off and show that DMEP provides a controllable mechanism for selecting different operating points under different deployment constraints.

\textbf{Impact of Pruning Timing.} 
DMEP delays structural pruning until the exploratory phase has progressed sufficiently, allowing routing patterns to become more stable. We ablate this design choice by altering the pruning step under the $\tau=0.10$ setting. Pruning too early (Step 50) prematurely terminates the exploration phase, resulting in a sparser architecture (26.7M parameters) that likely discards experts before they receive sufficient gradient updates, leading to a suboptimal accuracy of 91.23\%. Conversely, pruning too late (Step 250) retains more capacity (35.3M parameters), possibly because the model continues to allocate capacity under the influence of load balancing for longer than necessary, yielding no additional accuracy gain (91.23\%). Pruning at Step 100 (exactly at the end of the first epoch) retains 31.2M parameters and achieves the highest accuracy. This suggests that, in our setting, one full epoch of dense exploration provides a favorable pruning point for establishing stable routing preferences before structural compression.

\subsection{Structural Insights: Post-Pruning Architecture}

\begin{figure}[ht]
    \centering
    \includegraphics[width=\linewidth]{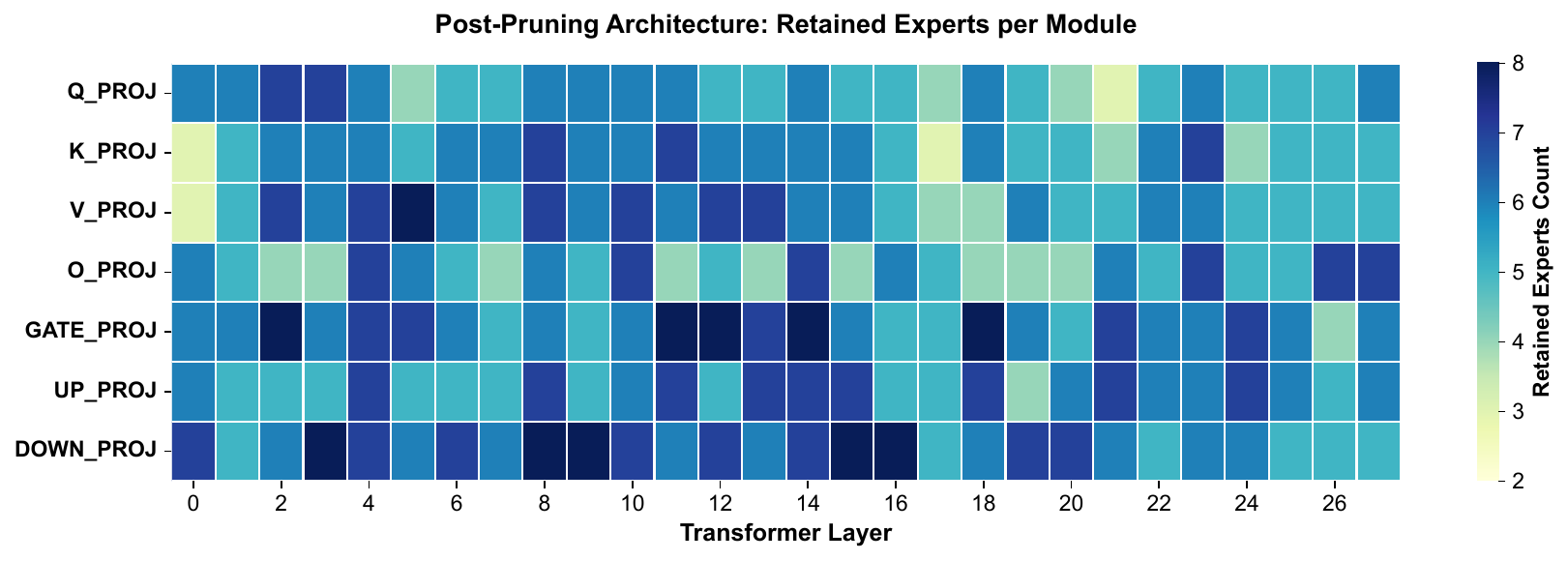}
    \caption{\textbf{Post-pruning Architecture (Retained Experts Heatmap).} The visualization maps the surviving experts across modules after DMEP ($\tau=0.10$). Attention projections (e.g., \texttt{Q\_PROJ}, \texttt{O\_PROJ}) exhibit higher sparsity due to skewed, specialized routing, whereas MLP projections (e.g., \texttt{GATE\_PROJ}, \texttt{DOWN\_PROJ}) retain broader capacity to support uniform knowledge distribution.}
    \label{fig:post_pruning_heatmap}
\end{figure}

To bridge our empirical results with the intrinsic routing behaviors, we visualize the exact architecture learned after DMEP is applied. \textbf{Fig.~\ref{fig:post_pruning_heatmap}} presents a heatmap of the retained expert count across different Transformer layers and modules for the Qwen3-0.6B model on ScienceQA ($\tau=0.10$). 

The visualization reveals a profound physical connection between routing heterogeneity and post-pruning capacity. According to our pruning criteria, experts are removed only if their utilization falls below the threshold $\tau$. Consequently, highly skewed modules—where routing is concentrated on a few specialized experts—naturally starve the remaining experts, leading to a sparser retained structure. Conversely, modules with uniform routing behavior distribute tokens evenly, keeping most experts above the survival threshold.

The heatmap perfectly reflects this dynamic. Attention projection modules (e.g., \texttt{Q\_PROJ}, \texttt{O\_PROJ}) appear distinctly lighter, frequently compressing down to 4 or 5 experts (averaging $\sim$5.3). This confirms that self-attention mechanisms exhibit highly specialized routing and can be aggressively compressed. In stark contrast, MLP modules (e.g., \texttt{GATE\_PROJ}, \texttt{DOWN\_PROJ}) remain densely populated, often retaining 7 or 8 experts (averaging $\sim$6.4). This aligns with the consensus that feed-forward networks act as broad knowledge repositories requiring extensive, uniformly distributed capacity. Ultimately, this demonstrates that DMEP does not blindly compress the network; instead, it autonomously uncovers the intrinsic dimensionality of each module, molding a rigid symmetric MoE into a highly adaptive, task-optimal asymmetric architecture.

%% file: content/related.tex
The scale of modern Large Language Models (LLMs) \cite{achiam2023gpt, touvron2023llama} makes full-parameter fine-tuning computationally prohibitive. Parameter-Efficient Fine-Tuning (PEFT) mitigates this by updating only a minimal set of introduced parameters \cite{houlsby2019parameter, li2021prefix}. Notably, Low-Rank Adaptation (LoRA) \cite{hu2022lora} has become the dominant paradigm by decomposing weight updates into trainable low-rank matrices. While variants like AdaLoRA \cite{zhang2023adalora} improve efficiency via dynamic rank allocation, the expressive power of LoRA remains fundamentally bounded by its low-rank constraint. For complex tasks requiring multi-step reasoning, a single low-rank projection often struggles to capture highly heterogeneous semantic features \cite{qing2024alphalora}.

To transcend dense capacity limits without proportionally increasing computational costs, recent works integrate Mixture-of-Experts (MoE) \cite{shazeer2017outrageously} with PEFT. Frameworks such as LoRAMoE \cite{dou2023loramoe}, MixLoRA \cite{li2024mixlora}, and MoELoRA \cite{liu2024moe} replace standard dense adapters with an ensemble of LoRA experts, using a gating network and an auxiliary load-balancing loss ($L_{\text{aux}}$) for conditional computation. However, these architectures typically rely on a static and symmetric structural prior, uniformly assigning an equal number of experts to every projection module across the network. This uniform allocation inherently ignores the functional heterogeneity of different Transformer components, inevitably leading to localized capacity over-provisioning and severe parameter redundancy \cite{fedus2022switch}.

To address this redundancy, recent research has explored asymmetric MoE allocation and compression. Methods like MoLA \cite{gao2024higher} and MALoRA \cite{wang2025malora} vary expert counts across layers, while DA-MoE \cite{aghdam2024moe} and LoRA-SMoE \cite{xu2025sensitivity} adjust routing based on sensitivity metrics. Despite these advances, existing asymmetric approaches exhibit three key limitations: they often operate at a coarse layer-level granularity that overlooks distinct intra-layer module demands, rely on computationally heavy offline profiling prior to training, and apply pseudo-sparsity via binary zero-masks that fail to reduce the actual optimizer memory footprint. In contrast, our AEP framework discovers the optimal architecture online via lightweight token tracking and employs structural physical slicing to permanently excise pruned experts from both the computational graph and optimizer states, translating theoretical sparsity into tangible efficiency gains.

%% file: content/conclusion.tex
In this work, we address the structural inefficiencies and optimization conflicts inherent in standard symmetric LoRA-MoE fine-tuning, where uniform expert allocation leads to capacity over-provisioning and persistent load-balancing regularization restricts task-driven specialization. To overcome these limitations, we propose Asymmetric Expert Pruning (AEP), a dynamic, module-level optimization framework that transitions the network from a dense initialization to a task-adaptive asymmetric architecture. By employing structural physical slicing based on online token tracking and subsequently disabling the auxiliary balancing loss, AEP permanently removes low-utility experts from the computational graph and allows surviving experts to achieve sharper specialization. Extensive experiments across multiple model scales demonstrate that AEP substantially reduces the active trainable parameter footprint by 35\% to 43\% and accelerates training throughput, all while maintaining competitive downstream reasoning accuracy. Ultimately, AEP shifts the accuracy-efficiency Pareto frontier to a more favorable regime, providing a scalable and highly effective solution for resource-constrained deployments.